\documentclass[conference]{IEEEtran}
\usepackage{times}

\usepackage{multicol}
\usepackage[bookmarks=true]{hyperref}
\usepackage{graphicx}
\usepackage{amsmath}
\usepackage{amssymb}
\usepackage{booktabs}
\usepackage{multirow}
\usepackage{stfloats}
\usepackage{algorithm}
\usepackage{algpseudocode}
\let\labelindent\relax
\usepackage{enumitem}
\usepackage{lipsum}
\usepackage{cellspace, tabularx}
\usepackage{adjustbox}% [demo]
\usepackage{dsfont}% [demo]
\usepackage[noadjust]{cite}

\algnewcommand{\IFOR}[1]{\State\algorithmicfor\ #1\ \algorithmicdo}
\algnewcommand{\ENDIFOR}{\unskip\ \algorithmicend }
\DeclareMathOperator*{\argmin}{arg\,min}
\usepackage{etoolbox}
\makeatletter
\patchcmd{\@makecaption}
  {\scshape}
  {}
  {}
  {}
\makeatother
\usepackage[
singlelinecheck=false % <-- important
]{caption}
\def\onedot{.}
\def\eg{\emph{e.g}\onedot}

 \def\vs{\emph{vs}\onedot}

\def\etal{\emph{et al}\onedot}
\usepackage{color}

\IEEEoverridecommandlockouts
\begin{document}

% paper title
\title{ACID: Action-Conditional Implicit Visual Dynamics for Deformable Object Manipulation}

\author{\authorblockN{Bokui Shen$^{1,*}$\thanks{$*$ Work done during an internship at NVIDIA Research.} 
Zhenyu Jiang$^{2}$
Christopher Choy$^{3}$  
Silvio Savarese$^{1}$ Leonidas J. Guibas$^{1}$ \\ Anima Anandkumar$^{3,4}$
Yuke Zhu$^{2,3}$}
\authorblockA{$^{1}$Stanford University~~~$^{2}$The University of Texas at Austin~~~$^{3}$NVIDIA~~~$^{4}$Caltech}}

\maketitle

\begin{abstract}
Manipulating volumetric deformable objects in the real world, like plush toys and pizza dough, bring substantial challenges due to infinite shape variations, non-rigid motions, and partial observability. We introduce ACID, an action-conditional visual dynamics model for volumetric deformable objects based on structured implicit neural representations. ACID integrates two new techniques: implicit representations for action-conditional dynamics and geodesics-based contrastive learning. To represent deformable dynamics from partial RGB-D observations, we learn implicit representations of occupancy and flow-based forward dynamics. To accurately identify state change under large non-rigid deformations, we learn a correspondence embedding field through a novel geodesics-based contrastive loss. To evaluate our approach, we develop a simulation framework for manipulating complex deformable shapes in realistic scenes and a benchmark containing over 17,000 action trajectories with six types of plush toys and 78 variants. Our model achieves the best performance in geometry, correspondence, and dynamics predictions over existing approaches. The ACID dynamics models are successfully employed to goal-conditioned deformable manipulation tasks, resulting in a 30\% increase in task success rate over the strongest baseline. Furthermore, we apply the simulation-trained ACID model directly to real-world objects and show success in manipulating them into target configurations. For more results and information, please visit \href{https://b0ku1.github.io/acid/}{https://b0ku1.github.io/acid/}.
\end{abstract}

\IEEEpeerreviewmaketitle

\section{Introduction}
\label{sec:intro}

\begin{figure}
\centering
\includegraphics[width=\columnwidth]{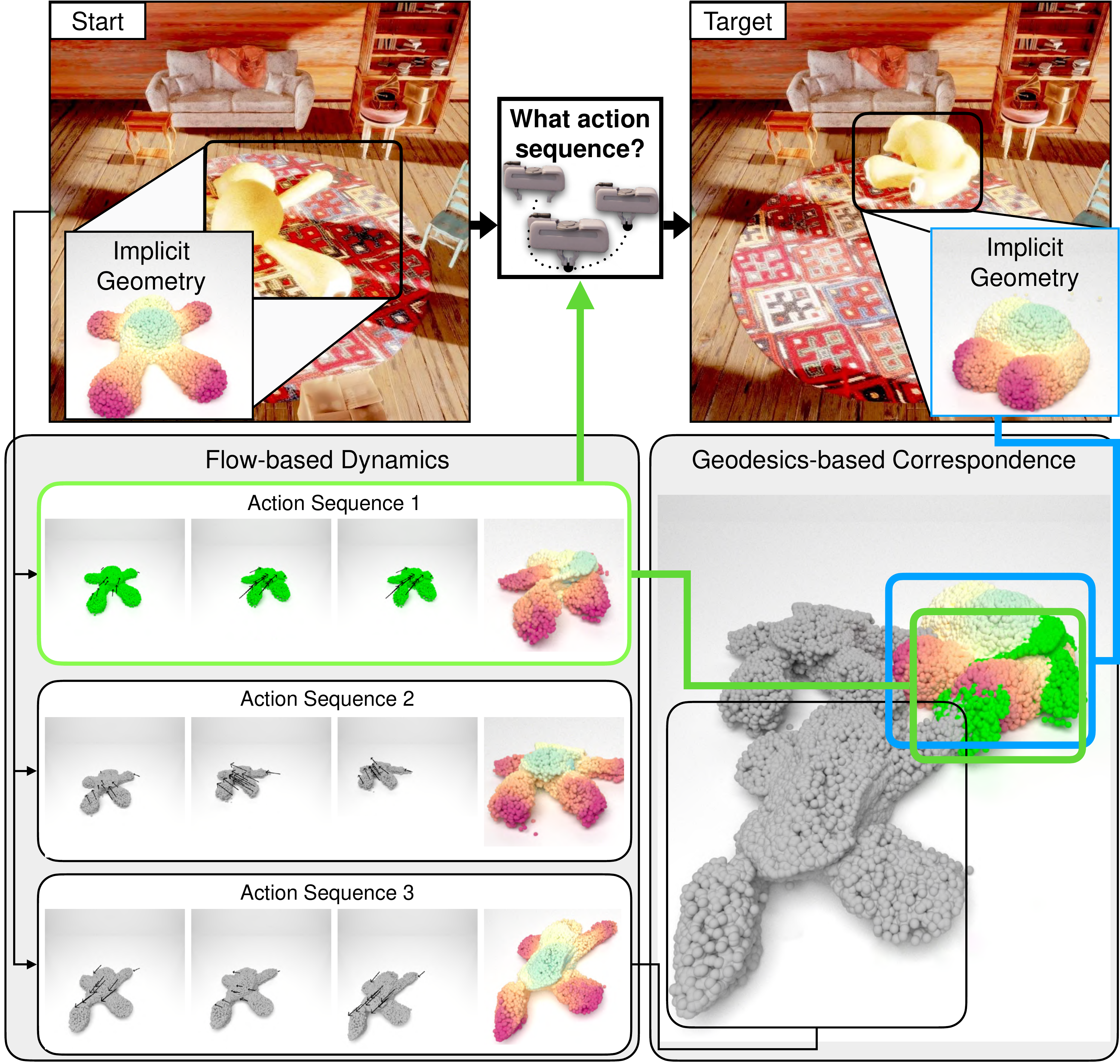}
  \caption{Perceiving deformable objects from raw visual signals and modeling their dynamics are challenging due to partial observations and complex non-rigid dynamics, making it difficult to determine what action sequence can manipulate an object into the desired configuration. We use implicit representations and geodesics-based correspondence learning to learn an end-to-end action-conditional visual dynamics model for realistic volumetric deformable objects manipulation.}
  \label{fig:pull}
\end{figure}

From playing with plush toys to making pizza dough, deformable objects are prevalent in many daily activities. Perceiving realistic deformable objects and modeling their dynamics over time play an integral role in building autonomous robots to interact with soft objects~\cite{sanchez2018robotic}.
However, estimating the geometry of deformable objects from raw visual signals and predicting their motions present significant challenges owing to their infinite continuous configuration spaces, complex non-linear dynamics, partial observability, and self-collision.

Decades of robotics research have developed dynamics models that characterize complex motions of non-rigid objects. Conventional approaches rely on high-fidelity mathematical models~\cite{jimenez2012survey,khalil2010dexterous,moore2007survey,saadat2002industrial} or recently their learned approximations~\cite{li2018learning,li2019propagation}.
These models, however, are hard to estimate from raw sensory data, hindering their applicability. In recent years, data-driven methods have demonstrated preliminary success in modeling 1D deformable linear objects like ropes~\cite{moll2006path,schulman2013tracking,roussel2015manipulation,tang2017state,yan2020self}, 2D flat deformable objects like cloth~\cite{miller2012geometric,li2014real,li2014recognition,mariolis2015pose,li2018model,lin2021learning}, or primitive 3D objects like sponges~\cite{navarro2014visual,petit2015real,navarro2016automatic,navarro2017fourier}. Nonetheless, existing methods struggle to handle volumetric deformable shapes, of which the 3D volumes and dynamics have to be inferred from partially observed visual data. Meanwhile, the 3D vision community has made great strides in modeling deformation from visual data. Specifically, implicit neural representations have shown great promise to model deformable objects and their dynamics with coordinate-based neural networks in high fidelity~\cite{niemeyer2019occupancy,jiang2020shapeflow,park2021nerfies,pumarola2021d}. Nonetheless, these methods focus on limited object categories and constrained motions rather than action-conditioned dynamics required for manipulation planning.

We introduce an action-conditional visual dynamics model for volumetric deformable objects manipulation in realistic 3D scenes, named ACID (\textbf{A}ction-\textbf{C}onditional \textbf{I}mplicit \textbf{D}ynamics). Our model learns joint representations for geometry, flow-based dynamics, and correspondence embedding fields end-to-end, as illustrated in  Fig.~\ref{fig:pull}.
Unlike explicit 3D representations (e.g. point cloud or voxels) widely used in dynamics modeling~\cite{byravan2017se3,tung20203d,xu2020learning,lin2021learning}, we use implicit neural representations to reconstruct high-fidelity full geometry and predict flow-based dynamics field from RGB-D observations. 
Moreover, measuring state changes accurately under non-rigid transformations benefits from establishing a dense correspondence over time. We propose a geodesic-aware contrastive learning loss to learn a correspondence embedding field to establish point-wise correspondence. The learned correspondence improves dynamics prediction and informs manipulation planning.

We develop a new simulation framework for manipulating deformable objects in realistic scenes using state-of-the-art physical simulation and photorealistic rendering techniques to train and evaluate ACID. Our framework is built with the PhysX engine in NVIDIA's Omniverse Kit. We use the framework to generate a  deformable visual dynamics dataset containing 17,000+ action trajectories with six types of plush toys of 78 different variants. 

Our model significantly outperforms existing dynamics models using explicit 3D representations, with a $23\%$ increase in shape mIoU and a $24\%$ decrease in dynamics mean-squared error. Our geodesics-based correspondence shows a $15\%$ increase in a matching metric. We use our model for model-based control of volumetric deformable object manipulation, resulting in $30\%$ increase in task success rate. Furthermore, we directly apply our model trained with simulated dynamics to manipulate physical objects with a robot arm. We show that our model successfully manipulates real-world plush toys of various shapes into target configurations.

\section{Related Work}
\label{sec:rw}
\begin{figure*}[h]
\centering
\includegraphics[width=\textwidth]{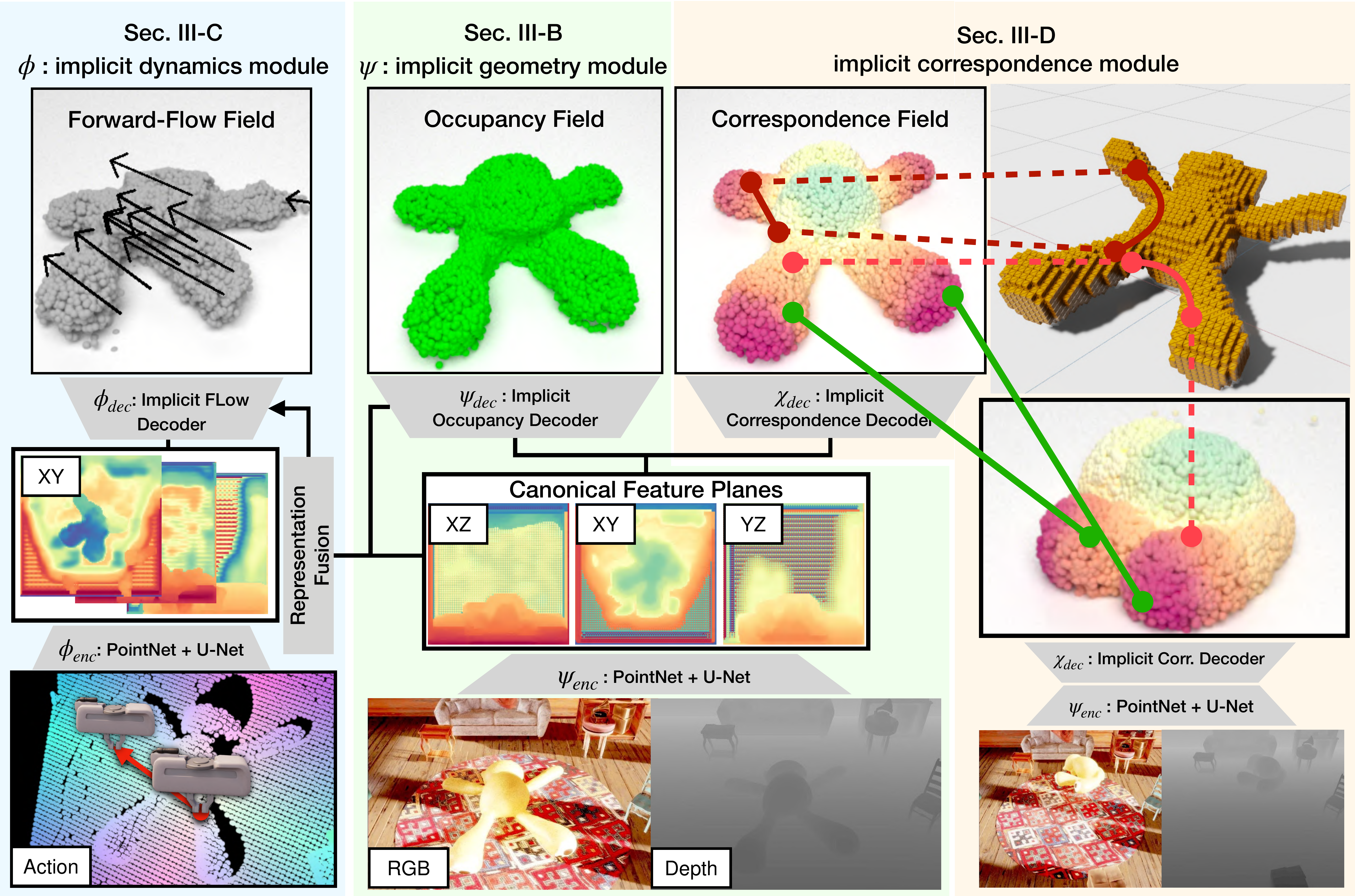}
  \caption{We introduce \textbf{ACID}, an end-to-end action-conditional visual dynamics model that jointly predicts geometry (Sec.~\ref{sec:3occ}), dynamics (Sec.~\ref{sec:3flow}), and correspondence (Sec.~\ref{sec:3geo}) as structured implicit representations. We extract state representation as explicit 3D shapes (point clouds) from implicit representations and visualize them here. First, the two encoders $\psi_{enc}$ for RGB-D inputs and $\phi_{enc}$ for robot actions produce three canonical feature planes, respectively. The Implicit Occupancy and Correspondence Decoders bilinearly interpolate the encoded feature planes at different 3D locations and generate: 1) an occupancy field and 2) a correspondence field trained with a geodesics-based contrastive loss. We illustrated the mechanism of contrastive learning on the upper right. We show here the t-SNE~\cite{van2008visualizing} color-coded features overlaid on an example pair. The Implicit Flow Decoder fuses the representations from the two encoders and predicts a one-step forward flow field.}
\label{fig:model}
\end{figure*}

\noindent \textbf{Learning Deformable Dynamics Models for Manipulation.} Robot manipulation of deformable objects has been a prolific field with wide-ranging applications in diverse industries. Various works have proposed approaches to extract  intermediate geometric representations of deformable objects and model their dynamics for manipulation. Such methods have shown success in 1D deformable linear objects (\eg, ropes) as line segments~\cite{moll2006path,schulman2013tracking,roussel2015manipulation,tang2017state,yan2020self}, 2D flat deformable objects (\eg, cloth) as a graph or a surface mesh~\cite{miller2012geometric,li2014real,li2014recognition,mariolis2015pose,li2018model,lin2021learning}. Others model simple 3D deformable objects (\eg, sponge) as their 2D contours~\cite{navarro2014visual,petit2015real,navarro2016automatic,navarro2017fourier}. Several comprehensive reviews \cite{sanchez2018robotic,jimenez2012survey} have captured the rich history in modeling deformable objects for robot manipulation. However, learning a full 3D representation in an end-to-end framework for realistic deformable objects is still beyond reach. 

Alternative approaches try to tackle deformable objects manipulation with learned dynamics models in pixel~\cite{finn2017deep,hafner2019learning,hoque2021visuospatial} or latent~\cite{yan2020learning,lippi2020latent,li2020causal,li20213d} spaces, rather than explicit physical computation. Pixel-space models directly predict future observations and measure prediction quality with the reconstruction loss or the perceptual loss. Latent-space models encode raw observations into latent state vectors (\eg, using self-supervision like reconstruction~\cite{lippi2020latent,li20213d}, contrastive loss~\cite{yan2020learning} or keypoints~\cite{li2020causal}), and learn the dynamics in the latent state space. These pixel-space and latent-space models can be learned end-to-end, but they do not explicitly represent the underlying 3D structures of deformable objects. In comparison, we introduce a new simulation framework with 3D supervision and study how 3D dynamics can be faithfully captured with structured representations. 

Other works from related fields developed learning models to approximate physics simulator~\cite{battaglia2016interaction,mrowca2018flexible,li2018learning,ummenhofer2019lagrangian,li2019propagation,sanchez2020learning}. With access to underlying physical state (\eg,~mass, velocity, acceleration), these works approximate the explicit calculations in physics simulator with learned models. However, these approaches are difficult to estimate from raw visual input.

\vspace{0.5mm}
\noindent \textbf{Implicit Representations of Geometry and Motion.}
Numerous works have demonstrated impressive performance in shape prediction using voxels~\cite{choy20163d, riegler2017octnet, murez2020atlas}, depth predictions~\cite{yao2018mvsnet, godard2017unsupervised}, point clouds~\cite{qi2017pointnet, qi2017pointnet++}, and meshes~\cite{kanazawa2018learning}. Recently, implicit representations~\cite{park2019deepsdf,mescheder2019occupancy,chen2019learning,sitzmann2019scene,sitzmann2020implicit,peng2020convolutional} have shown impressive performances representing shapes. Moreover, the implicit approach has shown impressive performance in beyond-3D tasks like spatial-temporal reconstruction~\cite{niemeyer2019occupancy,deng2020nasa} and shape deformation estimation~\cite{jiang2020shapeflow}. Prior works focus on rigid shapes from a large-scale shape repository~\cite{chang2015shapenet} or deformable shapes of a specific category (\eg,~humans)~\cite{loper2015smpl}. Our work focuses on extending these approaches to realistic deformable objects of different morphology (\eg,~plush snake vs. plush teddy), where geometry and motion must be learned jointly with actions. Furthermore, we consider deformable objects in realistic scenes where an agent must take obstacles into account for interaction with the object of interest.

\vspace{0.5mm}
\noindent \textbf{Geometric Correspondence Learning.}
Learning geometric correspondence has been widely studied~\cite{van2011survey,sahilliouglu2020recent}. We discuss two main branches of work. First is learning shape correspondences between deformable objects like human and animals~\cite{boscaini2016learning,litany2017deep,monti2017geometric,groueix20183d,halimi2019unsupervised,donati2020deep,eisenberger2020deep,yang2021continuous,eisenberger2021neuromorph}. These works take in surface meshes of two objects and learn their correspondences. In contrast to ours, these works learn correspondences between objects of the same category and morphology. Moreover, they need access to complete surface shapes, which are unavailable in real camera observations due to (self) occlusion. The second is learning dense features of point clouds for registration~\cite{deng2018ppfnet,choy2019fully,xie2020pointcontrast}. They used a partial point cloud for input, but these works have not been examined for deformable objects. 

\vspace{0.5mm}
\noindent \textbf{Deformable Object Simulation.}
Research in computer graphics has made huge progress in deformable object simulation~\cite{terzopoulos1988deformable,terzopoulos1991dynamic,muller2007position,nealen2006physically,moore2007survey}. Such progress has produced many robust physics simulation engines, and learning-based components have been incorporated for acceleration~\cite{macklin2014unified,erez2015simulation,hu2019taichi,hu2019difftaichi,hu2019chainqueen}. Such efforts have in turn powered recent simulators for deformable object manipulation~\cite{lin2020softgym,gan2020threedworld,huang2021plasticinelab}. By leveraging the GPU-accelerated finite element methods in Omniverse~\cite{nvidiaomniverse}, we extend existing simulators from primitive geometries to realistic objects in complex scenes.

\section{Action-Conditional Implicit Dynamics}
\label{sec:method}
In this section, we introduce ACID (Fig.~\ref{fig:model}), our action-conditional visual dynamics model for deformable object manipulation using structured implicit neural representations. We discuss the problem formulation and the three main components of ACID: geometry prediction, dynamics prediction, and correspondence learning.
Lastly, we discuss applying ACID to a deformable manipulation task using model-based planning. 

\subsection{Problem Formulation}
\label{sec:3prob}
We focus on manipulating deformable objects in clutter. Specifically, we build a robotic gripper controller that rearranges deformable objects from an initial position to the target position from RGB-D observations.
We denote the observation at time $t$ as $\mathbf{o}_t \in \mathcal{O}$, where $\mathcal{O}$ is the space of observations. We denote the action of a gripper at time $t$ as $\mathbf{a}_t = (p_{g}, p_{r}) \in \mathcal{A}$, where $\mathcal{A}$ is the space of actions. The action represents grasping a target object at point $p_{g}\in\mathbb{R}^3$, moving the gripper to $p_{r}\in\mathbb{R}^3$, and releasing the gripper.

We use a model-based approach. Thus, building a robust model of states and state transitions is of primal importance.
To this end, we learn state representations as both a full object geometry state $\mathbf{s}_t \in \mathcal{S}$ and a high-dimensional correspondence embedding space using neural networks. Here we denote $\mathcal{S}$ as the state space.
Learning this model requires three different functions: state representation, state transition, and state distance measure for planning.
We use neural networks for state representation $\psi:\mathcal{O}\rightarrow \mathcal{S}$ (Sec.~\ref{sec:3occ}) and state transition $\phi: \mathcal{S} \times \mathcal{A} \rightarrow \mathcal{S}$ (Sec.~\ref{sec:3flow}).
Lastly, $d(\cdot)$ is a state distance measure (Sec.~\ref{sec:3mani}). Conventional distance measures include non-parameterized metrics like mIoU or Chamfer distance, which do not take correspondences into account. Instead, we propose a learning-based correspondence module to compute more accurate distances under deformation (Sec.~\ref{sec:3geo}).

\subsection{Predicting Geometry from Partial Observation}
\label{sec:3occ}
To perceive deformable objects and reason about their dynamics in 3D space from partial observations, prior work has used various explicit 3D representations, among which partial point clouds~\cite{byravan2017se3,lin2021learning} and voxels~\cite{xu2020learning,tung20203d} are two of the most popular choices.
These representations, however, have to trade-off between memory footprint and geometric resolution crucial for accurate dynamics modeling.
Instead, we use an implicit geometry module $\psi:\mathcal{O} \rightarrow\mathcal{S}$ that takes a partial RGB-D observation $\mathbf{o}_t$ as input and performs both continuous implicit encoding as well as 3D completion.
The module contains two components: an observation encoder $\psi_{enc}:\mathcal{O}\rightarrow \mathbb{R}^{H\times W \times D}$ that encodes partial observation into a feature map and an Implicit Occupancy Decoder $\psi_{dec}: \mathbb{R}^3\times \mathbb{R}^D \rightarrow [0,1]$ that maps a coordinate and its queried feature into an occupancy probability.

We follow the architecture of Peng~\etal~\cite{peng2020convolutional} to create the observation encoder $\psi_{enc}$ that maps a point cloud to a set of 2D feature maps. We first back-project an RGB-D image into the 3D space of per-point RGB colors and use a shallow PointNet~\cite{qi2017pointnet} with local pooling to map the three-dimensional input point coordinates and their colors into a latent feature space. These point cloud features are then orthographically projected onto three canonical feature planes ($xy,xz,yz$ planes). Lastly, we pass these three canonical features maps of size $H\times W\times D$ through a U-Net~\cite{ronneberger2015u}.
Specifically, we use a hidden layer dimension of $D=64$ and U-Net of depth $4$ to generate three feature planes of size $128\times 128\times 64$.

These three feature planes encode the shape implicitly. To obtain the occupancy at each point in $\mathbb{R}^3$, we use bilinear interpolation to compute a local feature of this point from each of the canonical feature planes and pass the summed feature to a multi-layer perceptron (MLP) to extract occupancy. Technically, let $p \in \mathbb{R}^3$ be a query point and we project $p$ to three canonical planes and sum three features from the planes to get $\psi_{enc}(\mathbf{o}_t)|_p$. Then we pass it through $\psi_{dec}$, an MLP with skip connections, to predict the occupancy probability:
\begin{align}
    \psi_{dec}(p, \psi_{enc}(\mathbf{o}_t)|_p) \in [0, 1]
\end{align}
Given the implicit geometry module $\psi$, we can extract state $\mathbf{s}_t$ which is the explicit 3D shape of the object of interest. Specifically, we extract an interior 3D point cloud from the Implicit Occupancy Decoder by aggregating the coordinates whose occupancy probabilities are above a threshold $\tau$: 
\begin{align}
    \mathbf{s}_t = \{p \in \mathbb{R}^3| \psi_{dec}(p, \psi_{enc}(\mathbf{o}_t)|_p) > \tau\} \subseteq \mathbb{R}^3
    \label{eq:occupancy}
\end{align}

\subsection{Action-conditional Flow-based Dynamics Field}
\label{sec:3flow}
Using implicit representations allows us to model geometry beyond the grid resolution. Similarly, we adopt a coordinate-based neural model to model high-fidelity dynamics~\cite{niemeyer2019occupancy,jiang2020shapeflow}. We use an implicit dynamics module $\phi$ to model the action-conditioned dynamics $\phi: \mathcal{S} \times \mathcal{A} \rightarrow \mathcal{S}$. The implicit dynamics module consists of two components similar to the implicit geometry module: an encoder $\phi_{enc}:\mathcal{O}\times \mathcal{A} \rightarrow \mathbb{R}^{H\times W \times D}$ that encodes the partial observation and an action into a feature map, and the Implicit Flow Decoder $\phi_{dec}: \mathbb{R}^3\times \mathbb{R}^D \rightarrow \mathbb{R}^3$ that maps a coordinate and its queried feature into a one-step prediction of forward flow.

We encode the action jointly with the input partial observation using $\phi_{enc}$, with the same network architecture as our geometry module's encoder $\psi_{enc}$. Similar to geometry feature encoding, we first back-project the RGB-D image to a partial point cloud. Instead of using per-point RGB color as the point feature, here we fuse the action command with the partial point cloud into a point feature. Given an action $\mathbf{a}_t=(p_{g}, p_{r})$, for each $p_i$ in the partial point cloud, we calculate its distance to the grasp location and assign per-point feature as $(p_{g}-p_i, p_{r})$. This way we ensure that per-point features can capture the point's relative position to the action location. This featurized partial point cloud is similarly encoded into three canonical features planes of $128\times 128 \times 64$. 

For point $p \in \mathbb{R}^3$, we query the feature vector $\phi_{enc}(\mathbf{o}_t, \mathbf{a}_t)|_p$ in the same way as the Implicit Occupancy Decoder. Finally, the one-step forward flow is computed with the Implicit Flow Decoder $\phi_{dec}$: 
\begin{align}
    \phi_{dec}(p, \phi_{enc}(\mathbf{a}_t, \mathbf{o}_t)|_p) \in \mathbb{R}^3
\end{align}

\vspace{1mm}
\noindent \textbf{Representation fusion:}  We learn the implicit dynamics module $\phi$ jointly with the implicit geometry module $\psi$ (Sec.~\ref{sec:3occ}) through representation fusion. When predicting occupancy for point $p$, the encoder generates a per-point feature $\psi_{enc}(\mathbf{o}_t)|_p$ (Sec.~\ref{sec:3occ}). The Implicit Occupancy Decoder, which contains multiple fully-connected layers, further processes $\psi_{enc}(\mathbf{o}_t)|_p$ into a set of per-point intermediate representations, which we denote as $\mathbf{r}_p$ for the point $p$. We feed $\mathbf{r}_p$ and $\psi_{enc}(\mathbf{o}_t)|_p$ as additional inputs into the Implicit Flow Decoder. Consequently, the one-step forward flow prediction becomes:
\begin{align}
\phi_{dec}(p, \phi_{enc}(\mathbf{a}_t, \mathbf{o}_t)|_p, \psi_{enc}(\mathbf{o}_t)|_p, \mathbf{r}_p) \in \mathbb{R}^3
\end{align}

\subsection{Geodesics-based Contrastive Learning}
\label{sec:3geo}
When reasoning about a deformable object in varying configurations, a critical challenge is to establish dense correspondences between two visual observations of the same object in drastically different configurations. Prior work~\cite{choy2019fully,xie2020pointcontrast} has used contrastive learning for point cloud correspondence features. But they focus on 1) rigid objects, 2) static contrastive margin, and 3) Euclidean distance to define positive pairs. For deformable objects, we have to take deformation into account and the simple Euclidean distance is no longer accurate. For example, when a teddy bear's arm bends and touches its body, the arm and the body have zero Euclidean distance, but they belong to two separate parts that we must distinguish. Therefore, we propose to use the geodesic distance defined on the surface of a deformable object. It allows the arm and the body to have great distances even though they touch each other.
In this work, we use such a geodesic distance as the contrastive margin to learn more discriminative features.

For a pair of states $\mathbf{s}_t, \mathbf{s}_{t\prime} \subseteq \mathbb{R}^3$ under non-rigid deformation, we have a set of correspondences between the sets of points as $\mathcal{C}=\{(p,q)|p\in \mathbf{s}_t, q \in \mathbf{s}_{t\prime}\}$. In a contrastive learning setting~\cite{choy2019fully}, we can learn a point embedding $\mathbf{f}$ that for points $p,q$ minimize the loss:
\begin{align}\label{eq:contrloss}\begin{split}
L(\mathbf{f}_p,\mathbf{f}_q)= &I_{pq}[D(\mathbf{f}_p,\mathbf{f}_q)-m_{pos}]^2_+ \\ &+ \bar{I}_{pq}[m_{neg} - D(\mathbf{f}_p,\mathbf{f}_q)]^2_+\end{split}
\end{align} where $D(\cdot, \cdot)$ is a distance measure; $I_{pq}=1$ if $(p,q)\in \mathcal{C}$ and $0$ otherwise, $\bar{\cdot}$ is the NOT operator, $m_{pos},m_{neg}$ are margins for positive and negative pairs. To incorporate geodesics of the original shape manifold, we extend $L$ as:
\begin{align}\label{eq:geoloss}\begin{split}
    L_{geo}(\mathbf{f}_p,\mathbf{f}_q)=I^g_{pq}[D(\mathbf{f}_p,\mathbf{f}_q)&-m_{pos}]^2_+ \\ + \bar{I}^g_{pq} \big[ \log \left(\frac{d_{O}(p,q)}{d_{thres}}\right) &+ m_{neg} - D(\mathbf{f}_p,\mathbf{f}_q) \big]^2_+ 
\end{split}
\end{align} 
where $d_{O}(p,q)$ is the geodesic function described below, $d_{thres}$ is a geodesic threshold, and $I^g_{pq}=1$ if $d_{O}(p,q)< d_{thres}$ and $0$ otherwise.

We represent $\mathbf{f}$ as an embedding field in $\mathbb{R}^{32}$. The correspondence field is learned jointly with the occupancy prediction, sharing the same encoder $\psi_{enc}$. We use an Implicit Correspondence Decoder $\chi_{dec}$ to predict the point embedding: 
\begin{align}
\chi_{dec}(p, \psi_{enc}(\mathbf{o}_t)|_p) \rightarrow \mathbf{f}_p \in \mathbb{R}^{32}
\end{align}
To measure $d(\mathbf{s}_t,\mathbf{s}_{t\prime}) \rightarrow v \in \mathbb{R}$, we first extract a correspondence $\xi_{corr}$ between $\mathbf{s}_t$ and $\mathbf{s}_{t\prime}$ that minimize the aggregated feature distance among all possible matching:
\begin{align}\label{eq:corr_matching}
\xi_{corr} := \argmin_{\xi:\mathbf{s}_t \rightarrow \mathbf{s}_{t^\prime}} \sum_{p\in \mathbf{s}_t} D(\mathbf{f}_p,\mathbf{f}_{\xi(p)})
\end{align}
Given the matching $\xi_{corr}$, we can calculate the mean corresponded distance as: 
\begin{align}
d_{corr}(\mathbf{s}_t,\mathbf{s}_{t\prime},\xi_{corr})&:= \frac{1}{|\mathbf{s}_t|} \sum_{p\in \mathbf{s}_t} \Vert p - \xi_{corr}(p) \Vert^2
\label{eq:corr}
\end{align}
where $|\mathbf{s}_t|$ denotes the number of points in shape $\mathbf{s}_t$. 

\vspace{1mm}
\noindent \textbf{Geodesics computation:} For each volumetric deformable object $O$, we approximate it as a graph $G$ of connected tetrahedrons in a fixed resolution. For two arbitrary points $p,q \in O$, we can retrieve their corresponding tetrahedrons $t_p,t_q \in G$, and we can calculate the geodesic distance $d_O(p,q)$ between $p$ and $q$ as the shortest path distance between $t_p$ and $t_q$ in graph $G$. 
During non-rigid deformation, though each tetrahedron is deformed, the connectivity of the structure remains fixed, which establishes $d_O$ as a deformation-consistent metric.

\subsection{Planning with Dynamics Model}
We now explain how we use ACID for model-based manipulation planning following a target-driven setup, where the target is specified by an image of the goal state. Given the current and target configurations specified as RGB-D images $\mathbf{o}_0,\mathbf{o}_{target}$ respectively, model-based planning requires a cost function on a sequence of actions $\texttt{cost}(\mathbf{a}_{1},...,\mathbf{a}_{n})$, which guides the selection of actions that minimize the cost. Given the implicit geometry module, we first recover initial state $\mathbf{s}_0$ as the explicit full geometry at the current time using Eq.~\eqref{eq:occupancy} with the threshold $\tau=0.75$. Then, for state $\mathbf{s}_t$ at time $t$, we can estimate the one-step forward flow of each $p\in \mathbf{s}_t$ with the implicit dynamics module, described in Sec.~\ref{sec:3flow}:
\begin{align}\label{eq:planrollout}
    \mathbf{s}_{t+1}=\{p + \phi_{dec}(p, \phi_{enc}(\mathbf{a}_t, \mathbf{o}_t)|_p, \psi_{enc}(\mathbf{o}_t)|_p, \mathbf{r}_p) | p \in \mathbf{s}_t\}
\end{align}
Note that the function above takes as input $\mathbf{o}_t$, but for a future time step $t\neq 0$, the observation $\mathbf{o}_{t}$ is unavailable. We tackle this challenge as follows: First, we make the simplification that for a given point $p$, its features from the implicit geometry module $\psi$ ($\psi_{enc}(\mathbf{o}_t)|_p$ and $\mathbf{r}_p$) remain fixed across time steps. We make such approximation as point features are supervised with a correspondence loss, encouraging the same point to have a similar feature across different configurations. Thus we only need to evaluate $\psi$ once using $\mathbf{o}_0$ as input, to obtain $\psi_{enc}(\mathbf{o}_0)|_p$ and $\mathbf{r}_p$. Second, since $\phi_{enc}(\mathbf{a}_{t}, \mathbf{o}_{t})|_p$ does not depend on the per-point RGB feature (Sec.~\ref{sec:3flow}), we observe that we can approximate $\mathbf{o}_{t}$ without color by performing camera projection over the full object geometry $\mathbf{s}_{t}$. We then can randomly sample future actions from the projected point cloud of the object of interest. Thus, the calculation of Eq.~\eqref{eq:planrollout} can be performed iteratively to roll out future states under a given action sequence. 

To compute the cost with Eq.~\eqref{eq:corr}, we first recover the target state geometry as $\mathbf{s}_{target}$ with Eq.~\eqref{eq:occupancy}. We can establish correspondence $\xi_{corr}$ between $\mathbf{s}_0$ and $\mathbf{s}_{target}$ as in Eq.~\eqref{eq:corr_matching}. We then calculate the cost as the state distance between the roll-out state after $n$ actions $\mathbf{s}_{n}$ and the target state as:
\begin{align}
\texttt{cost}(\mathbf{a}_{1},...,\mathbf{a}_{n})=d_{corr}(\mathbf{s}_{n}, \mathbf{s}_{target},\xi_{corr})    
\end{align}
The action sequence of the lowest cost is chosen. 
\label{sec:3mani}

\subsection{Implementation Details}
In realistic scenes, the object of interest only occupies a small portion of the 3D space. At training time, we sample query points $p \in \mathbb{R}^3$ from a multivariate normal distribution with a mean of object's center-of-mass and with standard deviation proportional to the size of the object bounding box. The object of interest is specified via a 2D instance segmentation mask that is jointly passed into the network with RGB-D image. The Implicit Occupancy Decoder $\psi_{dec}$ is supervised by the binary cross-entropy loss between the predicted and the ground-truth value at the sampled 3D coordinates. The Implicit Flow Decoder $\phi_{dec}$ is supervised by the mean squared error. The Implicit Correspondence Decoder $\chi_{dec}$ is supervised with geodesic-based contrastive loss $L_{geo}$ in Eq.~\eqref{eq:geoloss}. 

\vspace{1mm}
\noindent \textbf{Decoder Architecture:} the decoders for occupancy, correspondence and flow all have a hidden layer dimension of $32$. For flow prediction, representation fusion is performed before each fully-connected layer. The intermediate features in the Implicit Flow Decoder are concatenated with the corresponding layer's feature from the Implicit Occupancy Decoder, and passed through a fully-connected layer.

\vspace{1mm}
\noindent \textbf{Training Procedures:} we use PyTorch~\cite{paszke2019pytorch} for model training and use the Adam optimizer~\cite{kingma2015adam} with a learning rate of $10^{-3}$ and weight decay of $10^{-4}$. All methods are trained for 36 epochs. We use the Adam optimizer~\cite{kingma2015adam} with a learning rate of $10^{-3}$ and weight decay of $10^{-4}$ for all methods. We perform evaluations on the validation set every 4,000 iterations and select the model with the lowest loss for dynamics prediction. We use a batch size of 12 for all baselines and all variants of our ACID model. Geodesic distances between pairs of points are pre-calculated prior to training. Training the ACID model takes 2 days on a TITAN RTX GPU.

\label{sec:3jimp}

\begin{figure}[t]
\centering
\includegraphics[width=\columnwidth]{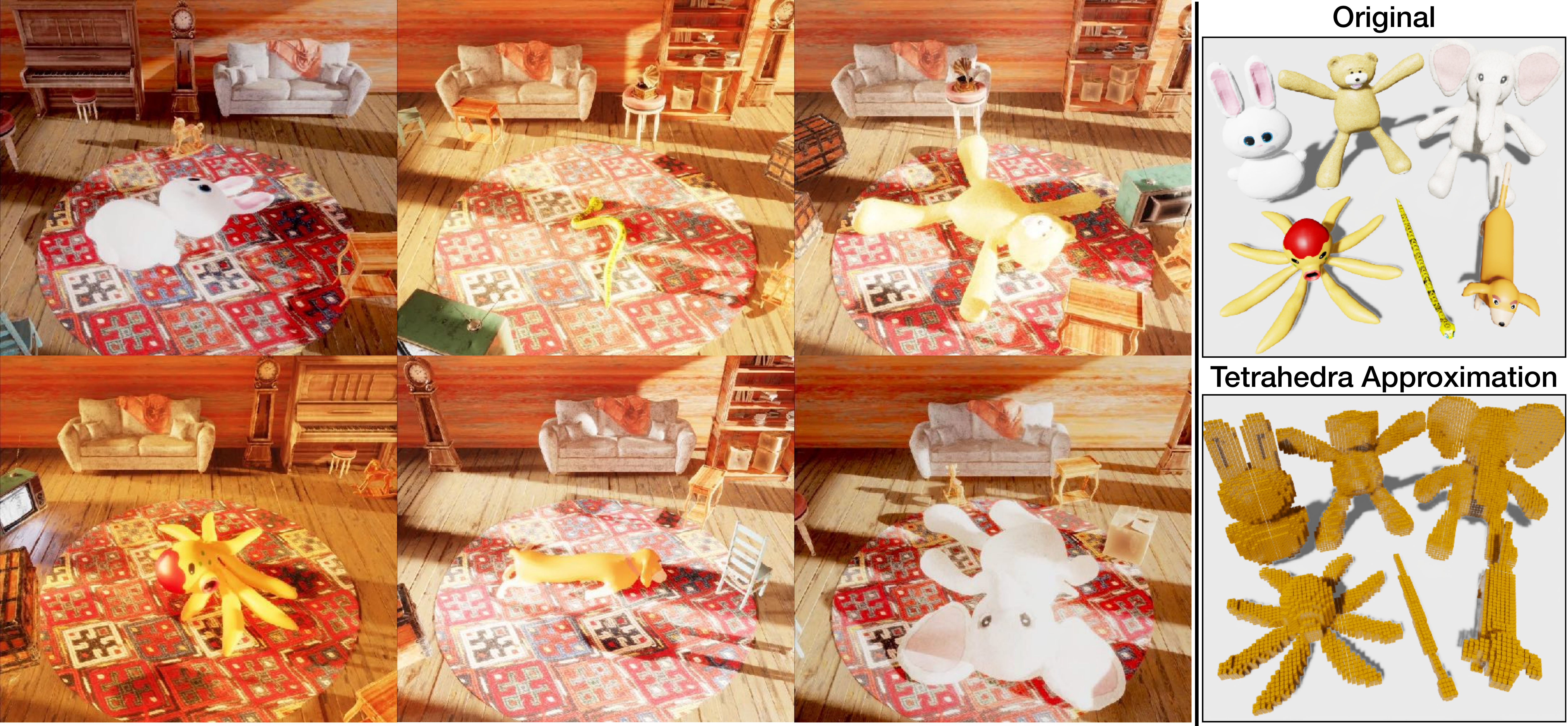}
  \caption{We introduce \textbf{PlushSim}, a manipulation environment of realistic volumetric deformable objects. Left: example images from the six types of plush toys, with lighting and layout randomization. Right: tetrahedral approximation used for finite element method simulation.}
  \label{fig:bench}
\end{figure}

\section{PlushSim Environment}
\label{sec:bench}

To quantitatively evaluate deformable objects manipulation in realistic scenes, we use a simulation environment where we can manipulate volumetric deformable objects and store the action information and the resulting ground-truth object motions. We develop a new framework that contains realistic actions and sensor simulation data for deformable plush toys in a visually and layout-wise randomized home scene.
\vspace{1mm}

\noindent \textbf{PlushSim Environment} Our framework is built with the PhysX engine in NVIDIA's Omniverse Kit. Omniverse features PhysX with GPU-accelerated finite element method (FEM) simulation that represents a deformable body volumetrically as a graph of connected tetrahedrons~\cite{nvidiaomniverse}. We use a fixed resolution for the object's tetrahedral approximation, ensuring consistent object simulation behavior. For physics simulation, we use a fixed time-step resolution of $\frac{1}{150}$ second, ensuring an accurate and consistent behavior for object-to-object and object-to-self collisions. On top of this engine, we created a realistic randomized attic scene and generated six types of plush toys with 78 variations for simulation. The attic scene contains 15 randomized furniture obstacles.  For each action sequence, the object pose, scene obstacles layout, and lighting are randomized to avoid overfitting (Fig.~\ref{fig:bench}). 

\begin{figure}[t]
\centering
\includegraphics[width=\columnwidth]{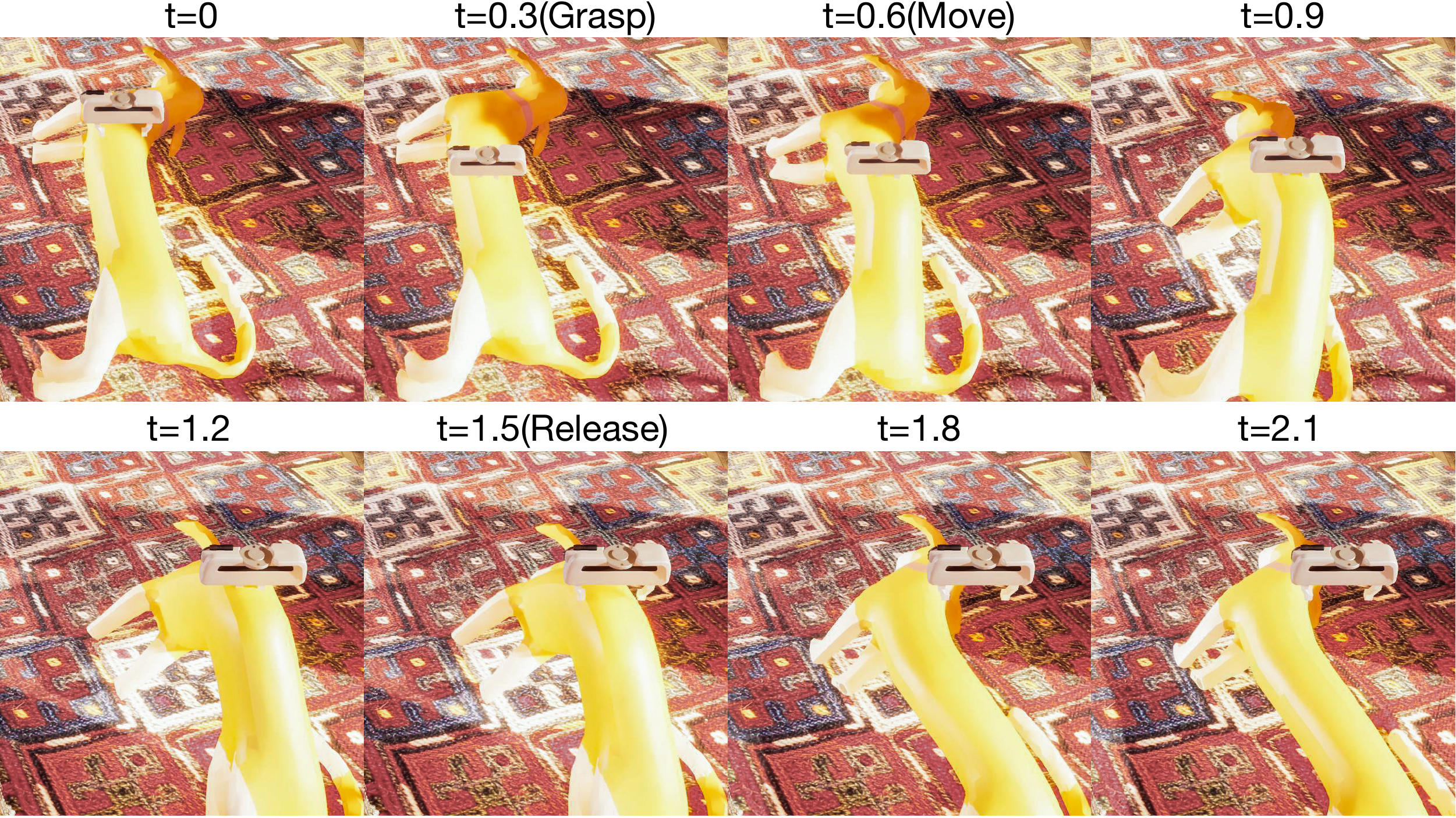}
  \caption{An example control command sequence. The sequence consists of \texttt{grasp}, \texttt{move}, and \texttt{release}. We visualize the scene every 0.3 seconds. We show more example sequences in the supplementary video.} 
\label{fig:example_act}
\end{figure}

We further implemented a set of manipulation API utilizing a Franka Emika Panda gripper over the simulation engine. The API contains parameterized control commands of \texttt{grasp}, \texttt{move}, and \texttt{release}. The \texttt{grasp} command takes the 3D location of the grasp center as input. The \texttt{move} command takes in a 3D displacement vector and moves the gripper in a straight line at a constant speed by the displacement. The \texttt{release} command takes no parameter. For collecting the dataset, we sample control commands as follows. For \texttt{grasp}, we sample uniformly among the object's visible points from the observations. For \texttt{move}, we sample a displacement vector in spherical coordinates $(r,\theta,\phi)$, where $r$ has a mean of 2.4 meters and a standard deviation of 0.8 meters; $\phi$ is uniform among $(0,2\pi]$; $\theta$ has a mean of $\frac{\pi}{4}$ and a standard deviation of $\frac{\pi}{6}$. The gripper is moving at a constant speed of 0.5 meters per second. We reset the scene and perform domain randomization after every 15 commands. We collect the data in parallel with 6 Titan RTX GPUs for 72 hours, resulting in a dataset of 17,665 actions. We show in Fig.~\ref{fig:example_act} an example action sequence where the gripper performs \texttt{grasp}, \texttt{move}, and \texttt{release}.

\vspace{1mm}

\noindent \textbf{Train/Test split.} For visual dynamics model evaluation, we use five types of plush toys for training and one for testing, which leads to 14,524 actions for training and 3,141 for testing. For the manipulation evaluation, we randomly sample 216 start and target configurations for rearrangement, each of which is equipped with 256 sampled action sequences.

\section{Experiments}
\label{sec:exp}
\begin{table}[t]
\centering
\begin{tabular}{lccccc}
\toprule
& \multicolumn{5}{c}{Action-Conditional Visual Dynamics} \\ \cmidrule(lr){2-6} & \multicolumn{2}{c}{Dynamics} & \multicolumn{2}{c}{Correspondence} & \multicolumn{1}{c}{Ranking} \\ \cmidrule(lr){2-3} \cmidrule(lr){4-5} \cmidrule(lr){6-6}
{Method} & vis\tiny{\textdownarrow} & full\tiny{\textdownarrow} & FMR\tiny{\textuparrow} & acc.\tiny{\textuparrow} & Kendall $\tau$\tiny{\textuparrow} \\ \hline
Pixel-flow~\cite{byravan2017se3} & 0.525 & - & - & - & -0.002 \\
Voxel-flow~\cite{xu2020learning} & 0.492 & 0.488 & - & - & 0.278 \\ \hline
ACID-No Corr & 0.434 & 0.385 & - & - & 0.534 \\
+Deep shells~\cite{eisenberger2020deep} & - & - & 0.085 & 0.035 & 0.516 \\ 
+FPFH~\cite{rusu2009fast} & - & - & 0.027 & 0.052 & 0.511 \\
+InfoNCE\cite{xie2020pointcontrast} & - & - & 0.081 & 0.061 & 0.510 \\
+FCGF~\cite{choy2019fully} & - & - & 0.665 & 0.204 & 0.526 \\ \hline
ACID-e2e-w/~\cite{xie2020pointcontrast} & 0.486 & 0.434 & 0.543 & 0.191 & 0.512 \\
ACID-e2e-w/~\cite{choy2019fully} & 0.486 & 0.436 & 0.751 & 0.240 & 0.527 \\ \hline
ACID-w/o fusion & 0.457 & 0.409 & 0.718 & 0.174 & 0.541 \\ \hline
ACID (Chamfer) & - & - & - & - & \textbf{0.547} \\ 
ACID (Ours) & \textbf{0.425} & \textbf{0.372} & \textbf{0.815} & \textbf{0.257} & 0.544 \\
\bottomrule
\end{tabular}
\caption{\textbf{Visual Dynamics Quantitative Comparison.} We report dynamics prediction, correspondence prediction, and action-ranking performances for baselines and our model in the test set. The results indicate that our implicit representations formulation consistently outperform existing approaches.}
\label{tab:big-table-split1}
\end{table}

We evaluate each component of our ACID model: geometry, dynamics, and correspondence learning. Specially, we aim at answering the following key questions:
\begin{itemize}[noitemsep,topsep=1pt,leftmargin=3mm]
\item (Sec.~\ref{sec:4eval_geomdyn}) does our model with implicit representations of geometry and dynamics outperform prior work using explicit representations?
\item (Sec.~\ref{sec:4eval_corr}) does our geodesics-based correspondence module outperform existing shape correspondence methods? 
\item (Sec.~\ref{sec:4eval_act}) does the high performance of each component translate to manipulation success? 
\end{itemize} 
Overall, we show that ACID outperforms baselines in all three components. Moreover, combining these components increases the task success rate in a volumetric deformable object manipulation task by 30\% over the strongest baseline. We further showcase directly applying the ACID model to manipulating real-world deformable objects. Finally, we discuss ablation studies and our limitations.

\begin{figure*}[t]
\centering
\includegraphics[width=\textwidth]{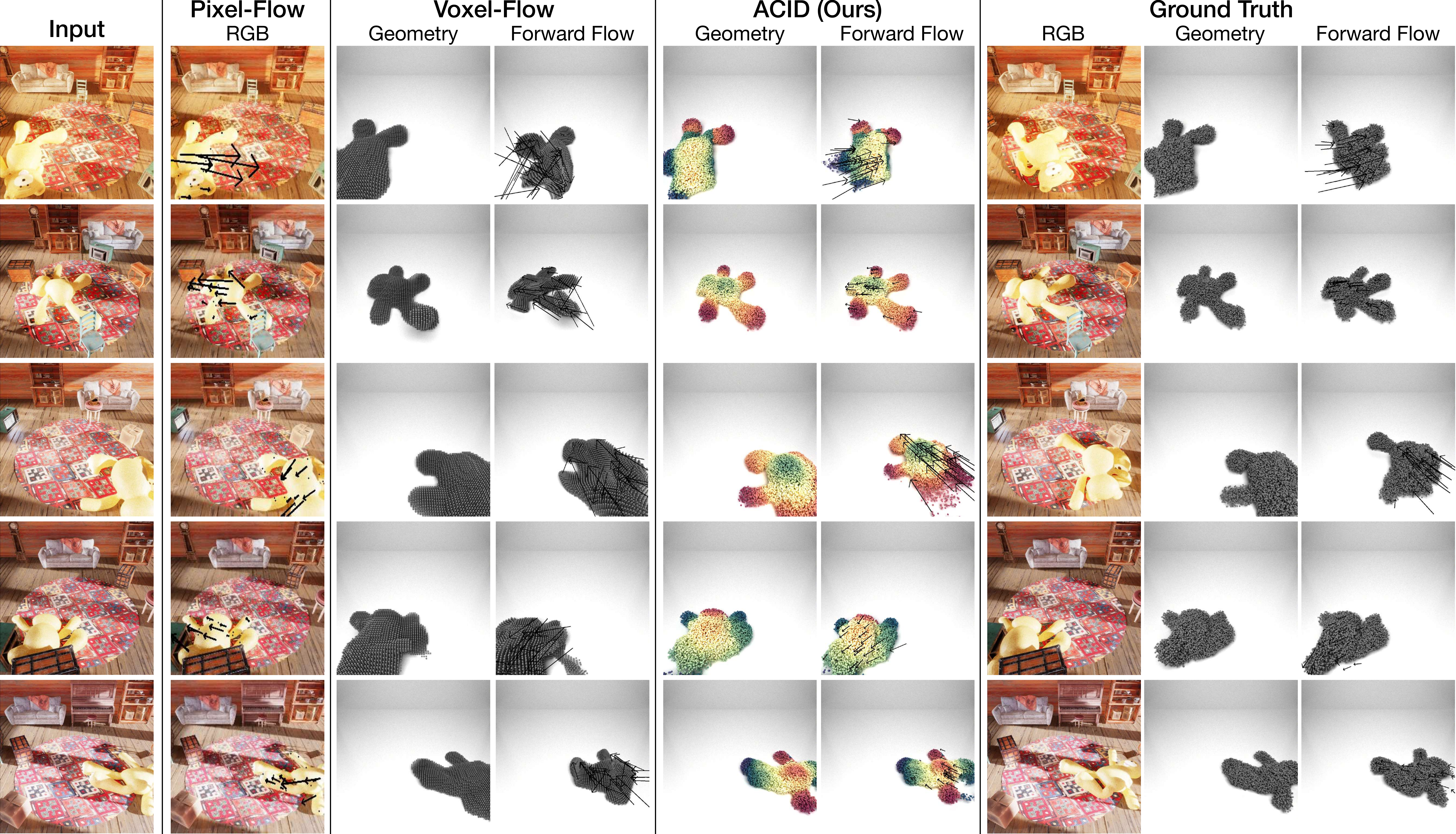}
  \caption{\textbf{Action-Conditional Visual Dynamics}. Qualitative comparison of ACID with \texttt{Pixel-Flow} and \texttt{Voxel-Flow} on the unseen toy type (bear). Flows are are visualized as black arrows. For ACID, we additionally show visualization of the t-SNE color-coded correspondence feature.}
\label{fig:exp_vis_qual}
\end{figure*}

\subsection{Geometry and Dynamics Evaluations}
\label{sec:4eval_geomdyn}
We conduct quantitative evaluations for geometry and dynamics prediction (Table~\ref{tab:big-table-split1}) and visualize qualitative results of different models in Fig.~\ref{fig:exp_vis_qual}.

\noindent \textbf{Geometry Evaluation:} We adapt the state-of-the-art voxel-based action-conditional visual dynamics models for rigid-body into a deformable setting as \texttt{Voxel-flow}~\cite{xu2020learning}, which represents the entire scene as a dense voxel grid of size $128\times 128\times 48$. The partial point cloud observation is encoded into a truncated signed distance function (TSDF), which is used to predict the instance mask of the object's full geometry and the scene flow of each pixel. To control the effect of jointly learning correspondences, we compare ACID model without contrastive supervision \texttt{ACID-No Corr}. We follow the standard procedure and report shape reconstruction quality with the mean Intersection over Union (mIoU) metric. \texttt{Voxel-flow} achieves a mIoU of $0.643$ in the test set, while \texttt{ACID-No Corr} achieves a mIoU of $0.796$. The results suggest that our implicit occupancy representation of geometry is more suitable for deformable shape reconstruction than previous explicit representation.

\noindent \textbf{Dynamics Evaluation:} Following prior work~\cite{xu2020learning}, we use the Mean Squared Error (MSE) to evaluate the predicted 3D scene flow and provide two metrics: MSE averaged over visible surfaces of the object of interest (\textit{vis}), and MSE averaged over all points with the object of interest (\textit{full}). We additionally compare with image-based approaches \texttt{Pixel-flow}, adapted from SE3-Nets~\cite{byravan2017se3}. As shown in Table~\ref{tab:big-table-split1}, \texttt{ACID-No Corr} achieves the best flow prediction performance of $0.434$ for visible surface and $0.385$ for full object. \texttt{Voxel-flow} performs significantly worse, with $0.492$ for \textit{vis} and $0.488$ for \textit{full}. This further suggests the usefulness of an implicit representation when modeling deformable objects and their dynamics. \texttt{Pixel-flow} performs the worst with $0.525$ MSE for \textit{vis}, corroborating that modeling deformable object dynamics benefits from reasoning over full object shape. 

Further, we observe that \texttt{ACID (Ours)}, which is trained jointly with correspondence, achieves better flow prediction performance of $0.425$ for \textit{vis} and $0.372$ for \textit{full}. By fusing the representations from the implicit correspondence decoder as well as the final embedding, the model can leverage auxiliary information for dynamics prediction. We ablate the effect of representation fusion by training ACID without representation fusion \texttt{ACID-w/o fusion}, which indeed achieves an inferior dynamics prediction of $0.457$ for \textit{vis} and $0.409$ for \textit{full}.

\subsection{Correspondence Evaluations}
\label{sec:4eval_corr}

We evaluate our geodesics-based correspondence against state-of-the-art shape correspondence baselines:
\begin{itemize}[noitemsep,topsep=1pt,leftmargin=3mm]
    \item \texttt{FPFH}~\cite{rusu2009fast}: uses an oriented histogram on pairwise geometric properties, which is widely adopted in robotics frameworks like ROS~\cite{quigley2009ros}.
    \item \texttt{InfoNCE}~\cite{xie2020pointcontrast}: state-of-the-art per-point dense feature learnt from a contrastive PointInfoNCE Loss.
    \item \texttt{FCGF}~\cite{choy2019fully}: convolutional contrastive geometric features that achieves state-of-the-art performances in registration.
    \item \texttt{Deep-Shells}~\cite{eisenberger2020deep}: state-of-the-art mesh-based correspondence model.
\end{itemize}

To evaluate the baselines, we used our model to perform shape reconstruction and retrieve the full object point clouds. For \texttt{Deep-Shells}, we use marching cube to generate the full mesh. We used author-provided pretrained weight for \texttt{Deep-Shells}. Following FCGF~\cite{choy2019fully}, we evaluate the feature-match recall with standard parameters~\cite{deng2018ppfnet}, which measures the percentage of matched pairs that can be recovered with high confidence. We additionally measure correspondence accuracy. A point is counted as matched accurately if its match is 5cm within the ground truth match. As indicated in Table~\ref{tab:big-table-split1}, our geodesics approach performs the best with an FMR of $0.815$ and a correspondence accuracy of $0.257$, significantly outperforms prior works. \texttt{FCGF} achieves the best performance amongst the baseline, with an FMR of $0.665$ and accuracy of $0.204$. 

We additionally examine different contrastive loss variations from prior works without geodesics~\cite{choy2019fully,xie2020pointcontrast}. The models are trained by only substituting the loss term for correspondence while keeping everything else unchanged. \texttt{ACID-e2e-w/}\cite{choy2019fully} and \texttt{ACID-e2e-w/}\cite{xie2020pointcontrast} perform worse than our geodesics-based contrastive loss. Interestingly, we no longer see the benefit of dynamics prediction for the ablated models. In fact, jointly training dynamics with correspondence~\cite{choy2019fully,xie2020pointcontrast} decreases dynamics prediction performance. It indicates that understanding geodesics plays a vital role in dynamics prediction. We made an interesting observation that jointly training dynamics and correspondence also improves correspondence prediction (\texttt{Ours} vs.\texttt{w/o-fusion}), further suggesting the synergies between correspondence and dynamics learning.

\begin{figure*}[t]
\centering
\includegraphics[width=\textwidth]{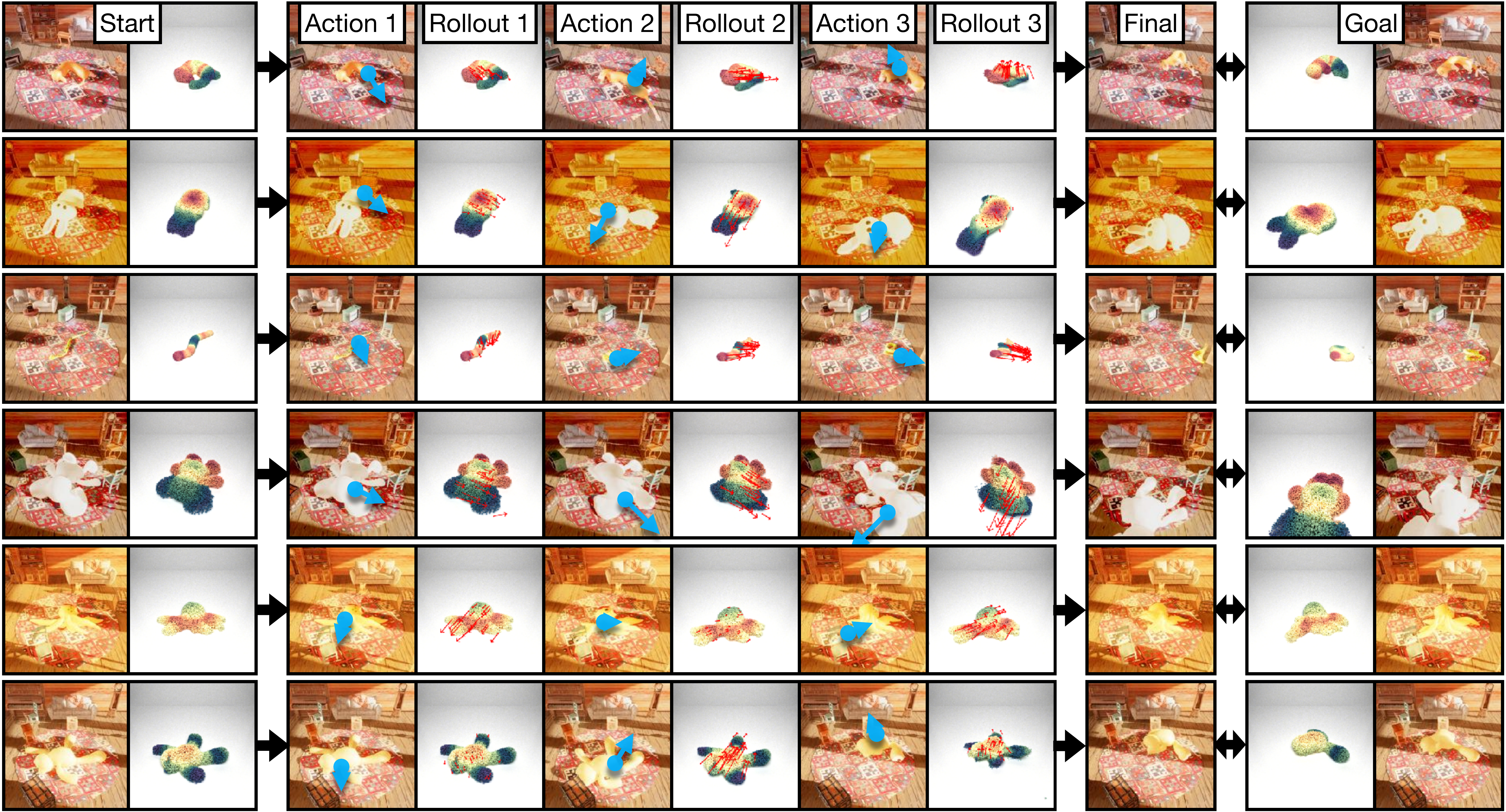}
  \caption{\textbf{Example action sequence roll-outs of our model:} Start and target configurations are specified as images. Future states are rolled out conditioned on action sequence. A cost measure selects the best action sequence, which is executed to get the final result.}
\label{fig:exp_act_qual}
\end{figure*}

\begin{table}[t]
\centering
\begin{tabular}{lccccc}
\toprule
 & \multicolumn{5}{c}{Manipulation Performance}\\ \cmidrule(lr){2-6} 
 & \multicolumn{4}{c}{Plan Execution Result} & \multirow{2}{*}{\begin{tabular}[c]{@{}c@{}}Success\\ Rate\tiny{\textuparrow}\end{tabular}} \\ \cmidrule(lr){2-5} 
{Method} & mIoU\tiny{\textuparrow} & F1\tiny{\textuparrow} & Chamfer\tiny{\textdownarrow} & $d_{corr}$\tiny{\textdownarrow} &  \\ \hline
Pixel-flow~\cite{byravan2017se3} & 0.247 & 0.435 & 1.210 & 1.720 & 19.4 \\
Voxel-flow~\cite{xu2020learning} & 0.281 & 0.495 & 1.005 & 1.459 & 22.2 \\ \hline
ACID-No Corr & 0.401 & 0.647 & 0.548 & 1.045 & 47.2 \\
+Deep shells~\cite{eisenberger2020deep} & 0.341 & 0.592 & 0.627 & 1.174 & 28.5\\ 
+FPFH~\cite{rusu2009fast} & 0.304 & 0.558 & 0.678 & 1.284 & 25.7 \\
+InfoNCE\cite{xie2020pointcontrast} & 0.314 & 0.566 & 0.660 & 1.268 & 28.5 \\
+FCGF~\cite{choy2019fully} & 0.382 & 0.628 & 0.580 & 1.082 & 45.7 \\ \hline
ACID-e2e-w/~\cite{xie2020pointcontrast} & 0.363 & 0.602 & 0.650 & 1.196 & 41.6 \\
ACID-e2e-w/~\cite{choy2019fully} & 0.330 & 0.574 & 0.676 & 1.273 & 30.5 \\ \hline
ACID-w/o fusion & 0.358 & 0.605 & 0.605 & 1.136 & 36.1 \\ \hline
ACID (Chamfer) & 0.435 & 0.701 & 0.451 & 0.947 & 54.3 \\ 
ACID (Ours) & \textbf{0.438} & \textbf{0.713} & \textbf{0.426} & \textbf{0.936} & \textbf{55.6} \\
\bottomrule
\end{tabular}
\caption{\textbf{Manipulation Quantitative Comparison.} We report the performances of plan execution and success rate in a manipulation task for baselines and variations of our model.}
\label{tab:big-table-split2}
\end{table}

\subsection{Manipulation Evaluations}
\label{sec:4eval_act}

\noindent \textbf{Task Setup: }we evaluate our models and all baselines in a volumetric deformable object manipulation task (Fig.~\ref{fig:exp_act_qual}). The scene consists of a deformable object of interest of a held-out type randomly posed in a randomized environment. The robot performs grasp-move-release actions as discussed in Sec.~\ref{sec:3prob}. The task is to rearrange the object of interest into the target configuration specified by an image. We define task success as the final configuration is within $0.5m$ from the target configuration.

Following prior work~\cite{xu2020learning}, for each start and target configuration, we sample 256 action sequences around the object of interest with a length of 3. We compute the cost for all action sequences and select the one with the lowest cost for each model. For models with correspondence prediction, the cost is measured as the corresponded distance $d_{corr}$ (Eq.~\eqref{eq:corr}) between rolled-out states and target states. For models without correspondence, the Chamfer distance is used for the cost. We further ablate our full model that was trained with correspondence prediction to use Chamfer distance as the cost measure \texttt{ACID (Chamfer)}.

\begin{table}[t]
\centering
\begin{tabular}{lcccccc}
\toprule
& \multicolumn{5}{c}{Trained Types} & Test \\ \cmidrule(lr){2-6} \cmidrule(lr){7-7}
Method 
& \adjustimage{width=.05\linewidth}{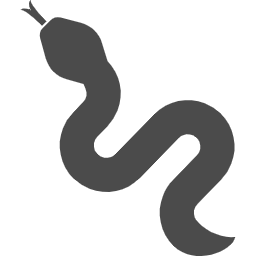} 
& \adjustimage{width=.05\linewidth}{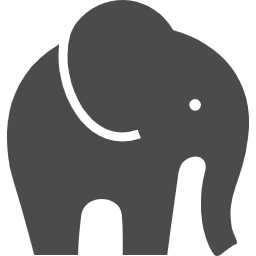} 
& \adjustimage{width=.05\linewidth}{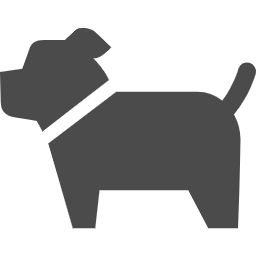} 
& \adjustimage{width=.05\linewidth}{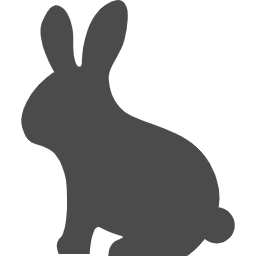} 
& \adjustimage{width=.05\linewidth}{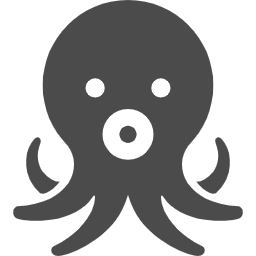} 
& \adjustimage{width=.05\linewidth}{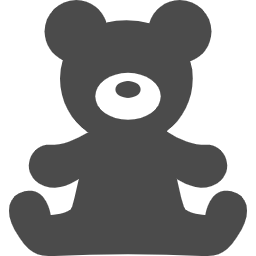} 
 \\ \hline
Pixel-flow~\cite{byravan2017se3} & 19.4 & 36.1 & 11.1 & 2.7 & 22.2 & 19.4 \\
Voxel-flow~\cite{xu2020learning}& 27.8 & 27.9 & 13.8 & 11.1 & 8.3 & 22.2 \\
ACID-No Corr & 34.3 & 45.7 & 14.2 & 11.4 & 31.4 & 36.1 \\
ACID (Ours) & \textbf{41.6} & \textbf{50.0} & \textbf{16.7} & \textbf{13.9} & \textbf{38.9} & \textbf{55.6} \\
\bottomrule
\end{tabular}
\caption{Manipulation task success rate by the toy type.}
\label{tab:train-test-table}
\end{table}

\vspace{1mm}
\noindent \textbf{Roll-out Ranking:} we first evaluate how accurately each model ranks the action sequences. We report the mean Kendall's $\tau$~\cite{kendall1938new} across 256 sampled sequences. It measures how well the predicted ranking agrees with the ground truth, with values close to 1 indicating strong agreement and -1 indicating strong disagreement. Variations of ACID model consistently show strong agreement with ground truth ranking, with an $0.26$ increase over \texttt{Voxel-Flow}.

\vspace{1mm}
\noindent \textbf{Plan Execution Results:} each model selects an action sequence based on the predicted lowest cost, we evaluate the selected action sequence's performance after being executed. We report mIoU, F-score~\cite{tatarchenko2019single} (F1), Chamfer distance, and $d_{corr}$ in Eq.~\ref{eq:corr} using the ground truth correspondence mapping $\xi_{gt}$ provided by the simulator. Table~\ref{tab:big-table-split2} shows that our model achieves the best execution results. As indicated by our performance over ablation baseline \textit{ACID (Chamfer)}, beyond the benefit of learning better feature representations, using $d_{corr}$ for action sequence cost measure also results in better model performance. Our model also achieves the best success rate, with a $30\%$ increase over the strongest existing visual dynamics model, \texttt{Voxel-Flow}. Visualization of example roll-outs are shown in Fig.~\ref{fig:exp_act_qual}.

\vspace{1mm}
\noindent \textbf{Per-Type Performance:}
We further report the planning success rate for five training toy types (see Table~\ref{tab:train-test-table}). As the table indicates, our model consistently outperforms prior approaches and self baseline. Interestingly, inter-type performances have a high variance, with smaller objects (dogs and rabbits) showing a lower success rate. It implies further challenges of modeling objects of various scales and reasoning at different resolutions. 

\begin{figure}
\centering
\includegraphics[width=\columnwidth]{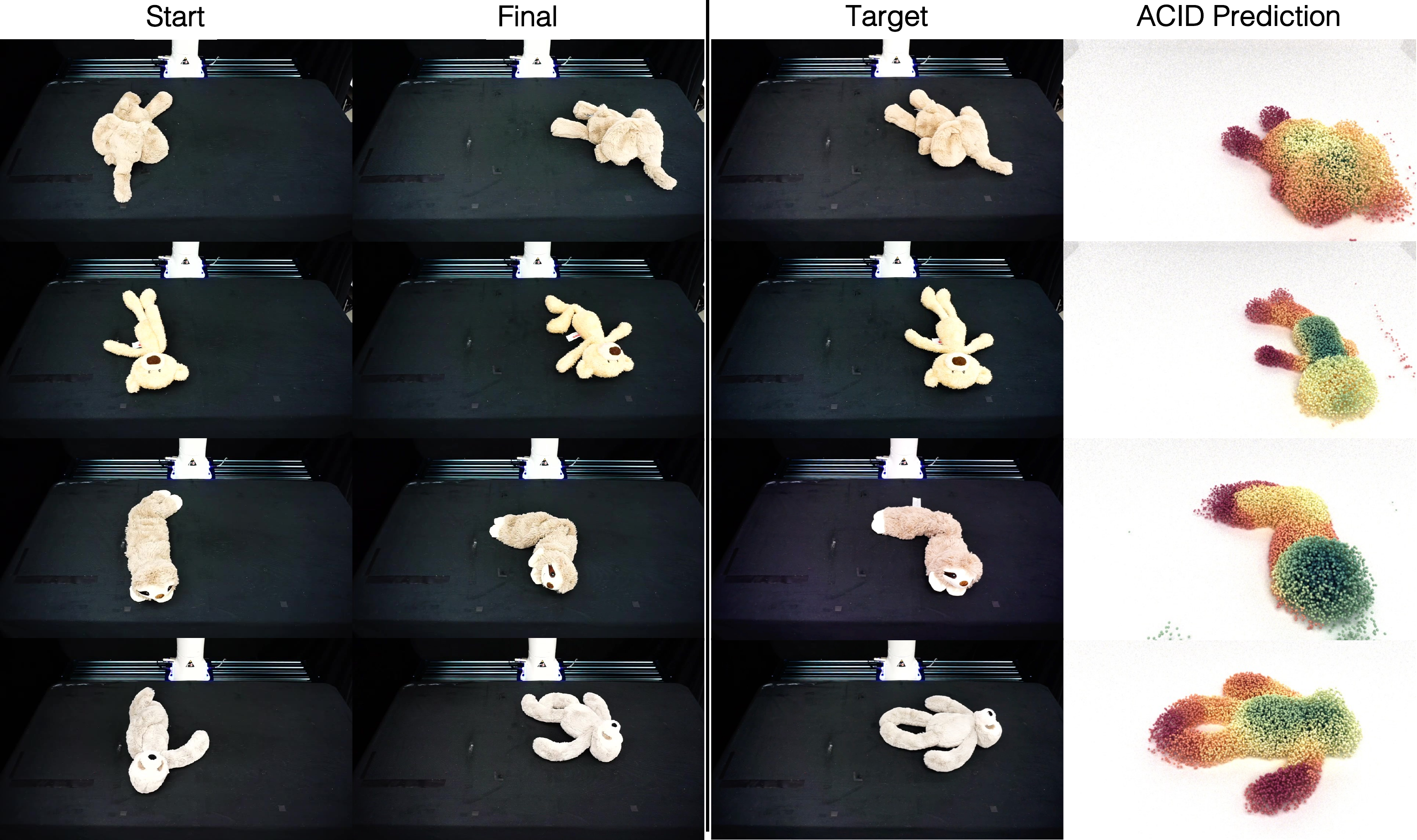}
  \caption{\textbf{Example manipulation results of our model on real world deformable objects:} left-most column shows the start configuration; second to left shows the final manipulation result. The right side shows the target configuration and the ACID predicted correspondence between start and target configuration.} 
\label{fig:real_act}
\end{figure}

\subsection{Real Robot Deployment}
\label{sec:4real}
We apply our ACID model trained with the PlushSim simulated dynamics to real-world objects. Similar to the setup in Sec.~\ref{sec:4eval_act}, we test the visual dynamics model on rearranging the object of interest into a target configuration specified by a goal image. We evaluated our model on 4 different plush animals. We use a 7-DoF Franka Emika Panda arm equipped with an Intel RealSense D435i camera mounted on the gripper. We sample 128 actions around the object of interest, among which we select the one with the lowest corresponded distance $d_{corr}$ computed with the ACID model. We set a maximum action horizon of 5. As shown in Fig.~\ref{fig:real_act}, our model correctly establishes the shape correspondence of real objects. Furthermore, it can successfully manipulate these plush animals into a pose similar to the target configuration. For video trajectories of the real robots, please visit our website.

\subsection{Discussion of Limitations}
\label{sec:4modellim}
Several limitations of ACID can be investigated and improved in future work.

First, although we build simulation environments of volumetric deformable objects with a high degree of visual and physical realism, applying ACID to real-world deformable objects requires careful engineering. Sim2real transfer of volumetric deformable object manipulation presents unique challenges compared to rigid bodies. First, object dynamics are higher dimensional and complex (flow fields \vs~SE(3)), which could enlarge the reality gap. Secondly, primitive actions like grasping for volumetric deformable object remains a challenging problem~\cite{huang2021defgraspsim}. Nonetheless, we believe recent advances in sim2real transfer methods for 1D and 2D deformable objects~\cite{wu2019learning,sundaresan2020learning,hoque2021visuospatial} could offer insight into closing the reality gap for volumetric deformable objects.

Second, since we randomly sample action trajectories, the dataset lacks variety towards the long-tail of fine-grained manipulation behaviors (\eg, twisting the toy snake into a loop). Manipulating object into a complex configuration is difficult for the current approach, and incorporating long-tail actions into training poses an exciting challenge for future work.

\section{Conclusion}
\label{sec:conclusion}
We introduce ACID, an action-conditional visual dynamics model for volumetric deformable objects manipulation based on structured implicit neural representations.
We proposed novel techniques for learning joint representations of dynamics field, occupancy field, and correspondence embedding field for action-conditional dynamics. Our results suggest that these techniques combined led to better generalization and higher manipulation performance. In a broader scope, these promising results shed light on how to build intelligent robots that can effectively reason about realistic deformable objects and their dynamics. One of the possible future directions is to model finer-grained dynamics and deploy the learned model on physical hardware.

\vspace{1mm}
\noindent \textbf{Acknowledgement:} We would like to sincerely thank Michelle Lu, Philipp Reist, and Cheng Low for help with Omniverse Kit simulation. We would like to sincerely thank De-An Huang, Linxi Fan, Zhiding Yu, the NVIDIA AI-ALGO team, other NVIDIA colleagues and colleagues from the Stanford Geometric Computation Group for the discussion and constructive suggestions.

%% Use plainnat to work nicely with natbib. 
\bibliographystyle{unsrt}
\bibliography{references}

\end{document}

% paper title
\title{Supplementary Material for \textit{ACID: Action-Conditional Implicit Visual Dynamics for Deformable Object Manipulation}}

\maketitle

\IEEEpeerreviewmaketitle

In this supplementary document, we provide detailed definitions for all evaluation metrics for geometry prediction, dynamics prediction, correspondence evaluation, ranking evaluation, and plan execution evaluation (Section~\ref{sec:metric}). Additional qualitative results for comparison with existing action-conditional visual dynamics models can be found in Section~\ref{sec:visviz}. Additional action sequence roll-outs for our ACID model can be found in Section~\ref{sec:rollviz}.

\section{Metrics}
\label{sec:metric}

In this section, we provide the formal definitions of the metrics that we use for evaluation. We define \texttt{mIoU} for geometry evaluation; \texttt{vis} and \texttt{full} mean-squared error for dynamics evaluation; \texttt{FMR} and \texttt{acc} for correspondence evaluation; \texttt{Kendall's} $\tau$ for ranking evaluation; \texttt{F-score} and \texttt{Chamfer} for planning execution evaluation. $d_{corr}$ and \texttt{success rate} have been defined in the main paper.

\subsection{Geometry Metric}
\noindent \textbf{mIoU:} we follow Peng \etal\cite{peng2020convolutional} for volumetric interaction over union calculation. Let $\mathcal{S}_{pred}$ and $\mathcal{S}_{GT}$ be the set of all points that are inside of the predicted and ground-truth shapes, respectively. The volumetric IoU is the volume of two shapes’ intersection divided by the volume of their union: 
\[
 \texttt{IoU}(\mathcal{S}_{pred}, \mathcal{S}_{GT}) = \frac{|\mathcal{S}_{pred} \cap \mathcal{S}_{GT}|}{|\mathcal{S}_{pred} \cup \mathcal{S}_{GT}|} 
\]
We randomly sample 100k points from the bounding boxes and determine if the points lie inside or outside $\mathcal{S}_{pred}$ and $\mathcal{S}_{GT}$, respectively. The \texttt{mIoU} is the mean volumetric intersection over union across test set.

\subsection{Dynamics Metric}
\noindent \textbf{vis and full MSE:} we follow Xu \etal~\cite{xu2020learning} for flow mean-squared error. Let $\mathcal{S}_{GT}$ be the set of all points that are inside of the ground-truth shapes, and $\mathcal{V}_{GT}$ be the set of all points that are visible in the camera for the ground-truth shapes. And let $f_p$ be the predicted flow $\in \mathbb{R}^3$, and $\hat{f_p}$ be the ground-truth flow $\in \mathbb{R}^3$. The \texttt{vis} and \texttt{full} mean-squared error is defined as:
\[
\texttt{vis}=\sum_{p\in\mathcal{V}_{GT}} \frac{(f_p - \hat{f_p})^2}{|\mathcal{V}_{GT}|} \quad
\texttt{full}=\sum_{p\in\mathcal{S}_{GT}} \frac{(f_p - \hat{f_p})^2}{|\mathcal{S}_{GT}|}
\]

\subsection{Correspondence Metric}
\noindent \textbf{FMR (Feature-match Recall):} we follow Deng \etal\cite{deng2018ppfnet} for feature-match recall calculation, which  measures the percentage of fragment pairs that is recovered with high confidence. Let $\xi_{gt}$ be the ground truth mapping and $\xi_{pred}$ be the predicted mapping. Mathematically, FMR is calculated as:
\[
R = \mathds{1}([\frac{1}{\mathcal{S}_{GT}}\sum_{p\in\mathcal{S}_{GT}} \mathds{1}(\Vert \xi_{gt}(p) - \xi_{pred}(p) \Vert < \tau_1) ] > \tau_2)
\]
And $R$ is averaged across test set. $\tau_1=0.1m$ inlier distance threshold and $\tau_2=0.05$ or $5\%$ is the inlier recall threshold, following prior work~\cite{deng2018ppfnet,choy2019fully}.

\noindent \textbf{acc. (accuracy):} is defined as the percentage of points that are correctly matched, where being correctly matched means that the predicted matched point is within $0.05m$ from the ground-truth matched point:
\[
\texttt{acc.} =\frac{1}{\mathcal{S}_{GT}}\sum_{p\in\mathcal{S}_{GT}} \mathds{1}(\Vert \xi_{gt}(p) - \xi_{pred}(p) \Vert < 0.05
\]

\subsection{Ranking Metric}
\noindent \textbf{Kendall's $\tau$:} we use \texttt{Kendall's $\tau$}~\cite{kendall1938new} for ranking evaluation, which has been examined in a computer vision context before~\cite{dwibedi2019temporal}. Kendall's $\tau$ is a measure of the correspondence between two rankings, with values close to 1 indicating strong agreement, and values close to -1 indicating strong disagreement. In our scenarios, we have a predicted ranking of sampled action sequence $r_{pred}$ and a ground-truth ranking of the sampled action sequence $r_{gt}$. For action sequence $i,j$, the quadruplet of rank indices $(r_{pred}(i), r_{pred}(j), r_{gt}(i), r_{gt}(j))$ is said to be concordant if $r_{pred}(i) < r_{pred}(j)$ and $r_{gt}(i)< r_{gt}(j)$ or $r_{pred}(i)> r_{pred}(j)$ and $r_{gt}(i) < r_{gt}(j)$. Otherwise it is said to be discordant. And \texttt{Kendall's $\tau$} is defined as:
\[
\texttt{Kendall's } \tau = \frac{P - Q}{P+Q}
\]
where $P$ is the number of concordant pairs, $Q$ the number of discordant pairs.

\subsection{Planning Execution Metric}
\noindent \textbf{Chamfer distance:} The Chamfer distance is widely used for point set distance calculation~\cite{fan2017point}. For two point sets $S_1, S_2 \subseteq \mathbb{R}^3$, the Chamfer distance is defined as:
\[
\texttt{chamfer}=\sum_{x\in S_1} \min_{y\in S_2} \Vert x-y \Vert_2^2 + \sum_{y\in S_2} \min_{x\in S_1} \Vert x-y \Vert_2^2
\]
For us, $S_1$ represents the object shape specified in the target configuration, $S_2$ represents the shape of the object after actions are executed.

\noindent \textbf{F-Score:} we follow Tatarchenko \etal\cite{tatarchenko2019single} for F-score calculation. \texttt{Recall} is defined as the number of points in the ground-truth shape that lie within a certain distance to the source shape. \texttt{Precision} counts the percentage of points on the source shape that lie within a certain distance to the ground truth. The F-Score is then defined as the harmonic mean between precision and recall:
\[\texttt{F-Score}=2 \cdot \frac{\text{Precision}\cdot \text{Recall}}{\text{Precision} + \text{Recall}}\]
For us, the source shape is the final shape of the object after the action sequence is executed.

\section{Action-Conditional Visual Dynamics Visualizations}
\label{sec:visviz}

\begin{figure*}[t]
\centering
\includegraphics[width=\textwidth]{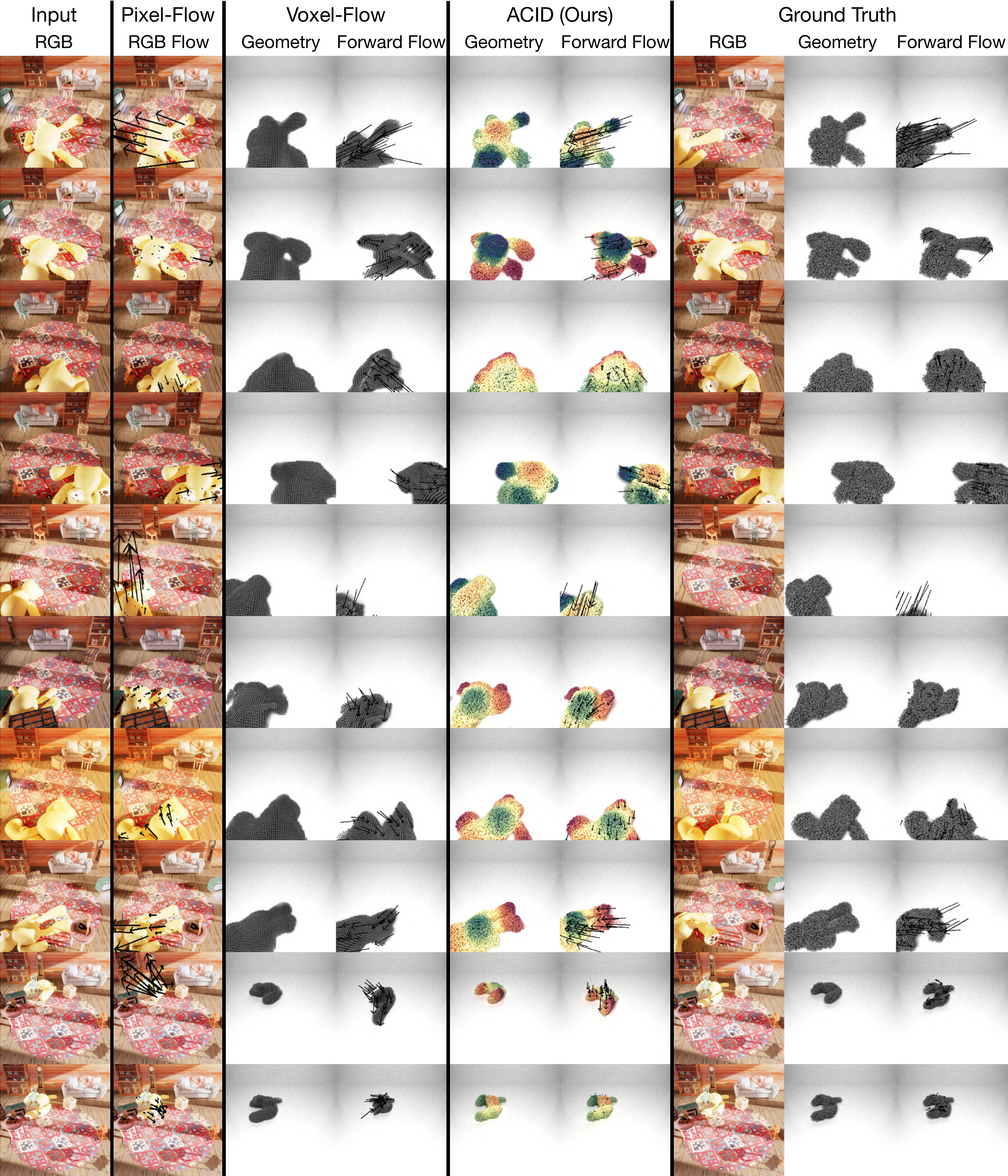}
  \caption{\textbf{Action-Conditional Visual Dynamics.} Comparison of ACID with \texttt{Pixel-Flow} and \texttt{Voxel-Flow} on unseen category.}
\clearpage\label{fig:exp_vis_qual1}
\end{figure*}

\begin{figure*}[t]
\centering
\includegraphics[width=\textwidth]{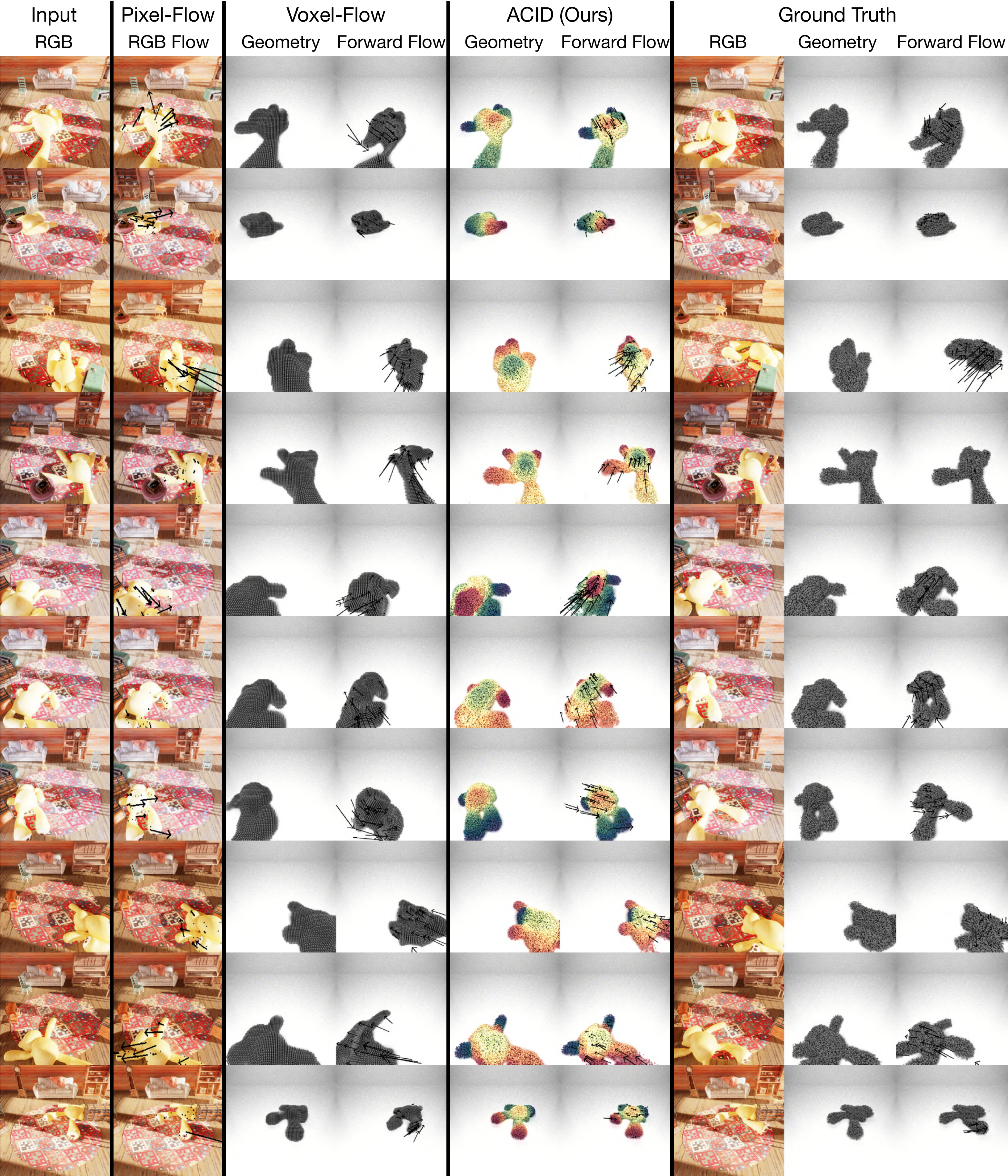}
  \caption{\textbf{Action-Conditional Visual Dynamics.} More Examples.}
\clearpage\label{fig:exp_vis_qual2}
\end{figure*}

We show here more example visualizations of our model comparing to the baselines \texttt{Pixel-Flow} and \texttt{Voxel-Flow}. As shown in Fig.~\ref{fig:exp_vis_qual1} and Fig.~\ref{fig:exp_vis_qual2}, our method consistently outperforms action-condition visual dynamics model baselines.

\section{Roll-outs Visualizations}
\label{sec:rollviz}
\begin{figure*}[t]
\centering
\includegraphics[width=\textwidth]{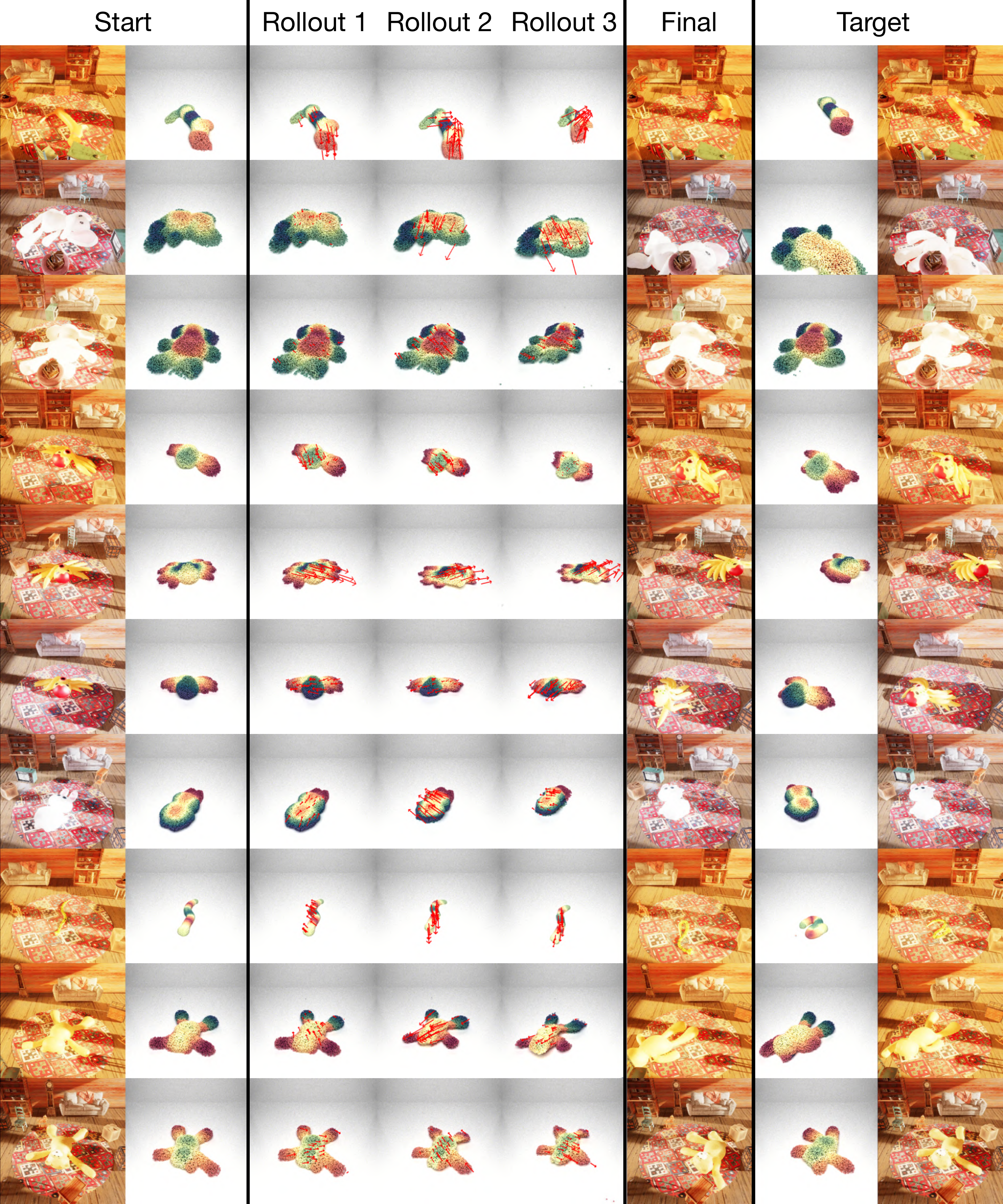}
  \caption{\textbf{Examples of the selected action sequence roll-outs of our model.} Start and target configurations are specified as images. Future states are predicted by our visual dynamics model conditioned on 256 sequences of 3-action commands. A cost measure selects the best action sequence to execute, of which the roll-out and final result are visualized in this figure.}
\clearpage\label{fig:exp_act_qual1}
\end{figure*}

We show here more example visualizations of ACID's planning results. We visualize the roll-outs of the selected action sequences with lowest cost under different start and target configurations. As shown in Fig.~\ref{fig:exp_act_qual1}, our method can estimate the dynamics of the object over multiple action commands and select reasonable action sequence.

\clearpage

%% Use plainnat to work nicely with natbib. 
\bibliographystyle{unsrt}
\bibliography{references}